\documentclass[11pt,twocolumn,twoside]{IEEEtran}
\usepackage{amsmath}
\usepackage[margin=0.75in,headheight=0.45in]{geometry}
\usepackage[pdftex]{epsfig}
\usepackage{amsfonts}
\usepackage{amssymb}
\usepackage{fancyhdr}
\include{graphicsx}
\begin{document}
\title{\vspace{0.2in}\sc DeepRain: ConvLSTM Network for Precipitation Prediction using Multichannel Radar Data}
\author{Seongchan Kim$^{1}$\thanks{Corresponding author: Sa-kwang Song, esmallj@kisti.re.kr $^1$Disaster Management HPC Technology Research Center, KISTI, Korea; $^2$Dept. of Big Data Science, UST-KISTI, Korea; Dept. of $^3$S\&T Information Science, UST-KISTI, Korea}, Seungkyun Hong$^{1,2}$, Minsu Joh$^{1,3}$, Sa-kwang Song$^{1,2}$}

\maketitle 
\thispagestyle{fancy}
\begin{abstract}
Accurate rainfall forecasting is critical because it has a great impact on people's social and economic activities. Recent trends on various literatures shows that Deep Learning (Neural Network) is a promising methodology to tackle many challenging tasks. In this study, we introduce a brand-new data-driven precipitation prediction model called \textit{DeepRain}. This model predicts the amount of rainfall from weather radar data, which is three-dimensional and four-channel data, using convolutional LSTM (ConvLSTM). ConvLSTM is a variant of LSTM (Long Short-Term Memory) containing a convolution operation inside the LSTM cell. For the experiment, we used radar reflectivity data for a two-year period whose input is in a time series format in units of 6 min divided into 15 records. The output is the predicted rainfall information for the input data. Experimental results show that two-stacked ConvLSTM reduced RMSE by 23.0\% compared to linear regression.
\end{abstract}

\section{Introduction}
Precipitation prediction is an essential task that has great influence on people's daily lives as well as businesses such as agriculture and construction. Acknowledging the importance of this task, meteorologists have been making great efforts to build advanced forecasting model of weather and climate, mainly focusing on Modeling \& Simulation based on HPC (High Performance Computing).

In recent years, studies using deep learning techniques have been drawing attention to improve prediction accuracy \cite{Shi2015, Zhang2017,SulagnaGopeSudeshnaSarkar2016,YongZhuang2016}. The Convolution Neural Networks (CNNs) and Recurrent Neural Networks (RNNs) are necessary techniques for the prediction of weather-related tasks. Several studies \cite{YongZhuang2016, Zhang2017} have employed each technique for precipitation prediction, and others \cite{Shi2015, SulagnaGopeSudeshnaSarkar2016} have tried using combinations of these. Primarily, convolutional LSTM (ConvLSTM), which is a variant of LSTM, was devised to embed the convolution operation inside the LSTM cell to model spatial data more accurately by Shi et al. \cite{Shi2015}. The authors demonstrated that ConvLSTM works on precipitation prediction in their experiments. However, they utilized three-dimensional and only one-channel data. In our study, we used the ConvLSTM for three-dimensional and four-channel data. 

\begin{figure}[h]
\centering
\includegraphics[width=0.9\columnwidth]{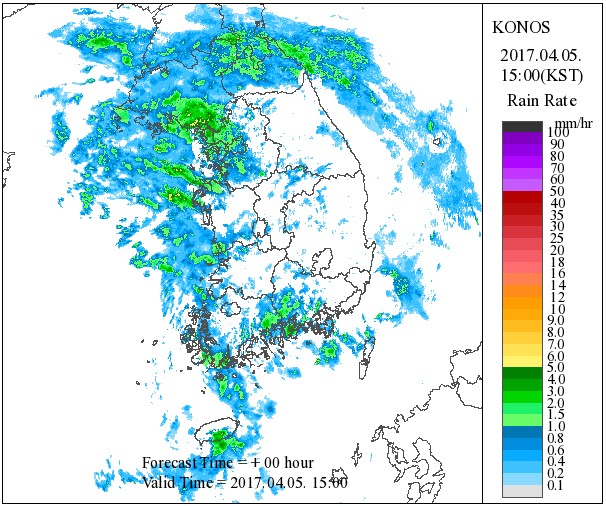}
\caption{Examples of radar map for rainfall rate}
\label{figure:radar}
\end{figure}

On the other hand, the data types 
used for rainfall-related prediction using the deep learning method are various. They include radar data \cite{Shi2015,Zhang2017}, past precipitation data, and atmospheric variable observation data such as temperature, wind, and humidity \cite{SulagnaGopeSudeshnaSarkar2016, YongZhuang2016}. In general, weather radar observations are the data used as inputs to numerical forecasting and hydrologic models to improve the accuracy of weather forecasts for hazardous weather such as heavy rains and typhoons \cite{KNWRC}. Specifically, weather radar data refers to data represented by a radar image that is composed using the moving speed, direction, and strength of a signal transmitted by a radar transmitter into the atmosphere and received after it has collided with water vapor or the like. For example, Figure \ref{figure:radar} shows a radar image of the Korean peninsula at 15:00 on April 5, 2017, and shows the rainfall rate in different colors depending on the degree of reflection.

In this study, to estimate the rainfall amount based on the weather radar data, we introduce \textit{DeepRain}, which applies ConvLSTM, one of the variants of LSTM. Our radar data is three-dimensional data (width, height, and depth) consisting of four channels (depth) from four altitudes. 
The contributions of this study are as follows:
\begin{enumerate}	
\item We adopt ConvLSTM first for three-dimensional and four-channel radar data to predict the rainfall amount.
\item We stacked the ConvLSTM cells for performance enhancement.
\item It was confirmed that the proposed method is more effective for predicting rainfall than linear regression and fully connected LSTM (FC-LSTM).
\end{enumerate}

This paper presents the rainfall forecasting model proposed in Chapter 3 and related research in Chapter 2. Section 4 shows the experimental procedure and results, and the paper concludes in Section 5.

\section{Related Work}
Zhuang and Ding \cite{YongZhuang2016} designed a spatiotemporal CNN to predict heavy precipitation clusters from a collection comprising historical meteorological data across 62 years. They used two-convolution, pooling, and fully connected layers. Zhang et al. \cite{Zhang2017} proposed a 3D-cube successive convolution network for detecting heavy rain. In this study, they cast rainfall detection as a classification problem and identified the presence of heavy rainfall by using radar data of several channels as an input to the convolution network. Gope et al. \cite{SulagnaGopeSudeshnaSarkar2016} proposed a hybrid method combining CNN and LSTM. Their model used outputs of CNN as inputs for LSTM. In their model, CNN and LSTM were considered as independent steps. For data, they employed atmospheric variables such as temperature and sea-level pressure. However, they did not consider radar data. Shi et al. \cite{Shi2015} devised a ConvLSTM model that enhanced an FC-LSTM model by replacing a fully connected layer with a convolution layer. However, that study was an attempt to generate a radar map for the future based on a past radar map using an many-to-many and end-to-end approach, while in our study, rainfall is forecast using a many-to-one (one-step time series forecasting) approach.

\section{Data}
The radar rainfall data used in the experiments were distributed by the Shenzhen Meteorological Administration in China for research purposes \cite{ShenzhenMeteorologicalBureau-Alibab}. The data, which were normalized and anonymized (Details about pre-processing of the data were not publicized.), were radar observations in the Shenzhen area. The data consists of numerical integer values (dBZ). There are 101 * 101 radar reflection values, each representing one cell after modeling a specific area of Shenzhen in grid form (101 * 101 $km^2$). The 101 * 101 numerical values are grouped into four groups (from an altitude of 3.5 km; 1-km intervals) and 15 intervals (every 6 min) for each altitude. (See Figure \ref{figure:DeepRain}.) That is, a total of 612,060 (=101 * 101 * 4 * 15) numerical values are listed. The ground truth is the measured rainfall amount ($mm^3$) from 1 h to 2 h in the area corresponding to the target area of 50 * 50 $km^2$ from the center of the grid. Therefore, one row of the data set is composed of 612,060 integer values (radar reflectivities) and one float value (ground truth). The complete data set consists of 10,000 rows randomly selected during a two-year period. We randomly divided the data into training (90\%), validation (5\%), and test data (5\%).

\section{Method}
ConvLSTM was introduced as a variant of LSTM by Shi in 2015 \cite{Shi2015} and it is designed to learn spatial information in the dataset. The main difference between ConvLSTM and FC-LSTM is the number of input dimensions. As FC-LSTM input data is one-dimensional, it is not suitable for spatial sequence data such as our radar data set. ConvLSTM is designed for 3-D data as its input. Further, it replaces matrix multiplication with convolution operation at each gate in the LSTM cell. By doing so, it captures underlying spatial features by convolution operations in multiple-dimensional data. The equations of the gates (input, forget, and output) in ConvLSTM are as follows:

\begin{equation}
i_{t}=\sigma (W_{xi}*x_{t}+W_{hi}*h_{t-1}+b_{i})\\
\end{equation}
\begin{equation}
f_{t}=\sigma (W_{xf}*x_{t}+W_{hf}*h_{t-1}+b_{f})\\
\end{equation}
\begin{equation}
o_{t}=\sigma (W_{xo}*x_{t}+W_{ho}*h_{t-1}+b_{o})\\
\end{equation}
\begin{equation}
C_{t}=f_{t}\circ C_{t-1}+\tanh(W_{xc}*x_{t}+W_{hc}*h_{t-1}+b_{c})\\
\end{equation}
\begin{equation}
H_{t}=o-{t}\circ\tanh(c_{t})
\end{equation}

where $i_{t}$, $f_{t}$, and $o_{t}$ are input, forget, and output gate. $W$ is the weight matrix, $x_{t}$ is the current input data, $h_{t-1}$ is previous hidden output, and $C_{t}$ is the cell state. The difference between equations in LSTM is that the convolution operation $(*)$ is substituted for matrix multiplication between $W$ and $x_{t}$, $h_{t-1}$ in each gate. By doing this, a fully connected layer is replaced with a convolutional layer, and then the number of weight parameters in the model can be significantly reduced.

\begin{figure}[h]
\centering
\includegraphics[width=1.0\columnwidth]{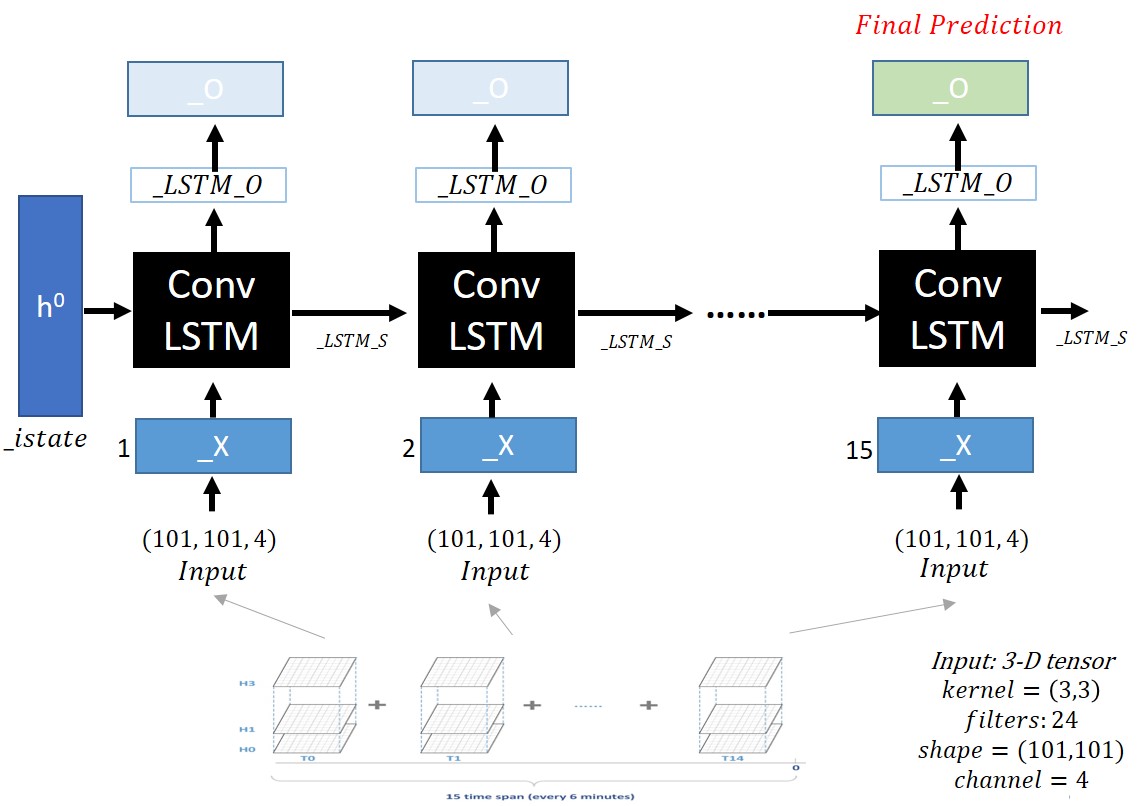}
\caption{\textit{DeepRain} architecture using ConvLSTM}
\label{figure:DeepRain}
\end{figure}

In this study, the problem involves predicting rainfall using test radar data with training radar data and its label (in actual fact, the measured amount of rainfall) on a large scale. We solve the problem by utilizing convLSTM. The structure of \textit{DeepRain} using convLSTM is shown in Figure \ref{figure:DeepRain}. The input data X of the model receives 15 items of data according to the time interval, and the input data of each node is 40,804 (which is reshaped as 101 * 101 * 4; three dimensions with four channels) for radar reflection value (integer) data. The output is the generated value $_O$ of the output gate of the last cell, which is the expected amount of rainfall for the input data. This model configuration (many-to-one) is a result of the data set which has the ground truth label is given to the radar data as a number of rainfall amount falling between 1 hour and 2 hour. Note that our model does not predict next sequences of labeling (many-to-many).

\section{Experiment}
We trained the proposed model (Figure \ref{figure:DeepRain}) with the Adam optimizer at a learning rate of 0.001. The input data (originally in txt format at 16 Gb) was transformed into a binary tfrecord (6.4 Gb) to improve the learning speed. We used random shuffling mini-batches for learning; the mini-batch size was set to 30. The training epoch used 50, and it took about 12 h to learn (one-stack ConvLSTM). The Root Mean Square Error (RMSE) was used to measure the prediction accuracy. The testbed environment configuration was as follows:
\begin{itemize}
    \item CPU: Intel R Xeon R E5-2660v3 @ 2.60GHz
    \item RAM: 128GB DDR4-2133 ECC-REG
    \item GPU: NVIDIA R TeslaTM K40m 12GB @ 875MHz (Dual)
    \item HDD: 4TB 7.2K RPM NLSAS 512n 12Gbps
    \item Framework: TensorFlow 1.2, Python 3.5.2
\end{itemize}

\begin{figure}[h]
\centering
\includegraphics[width=0.9\columnwidth]{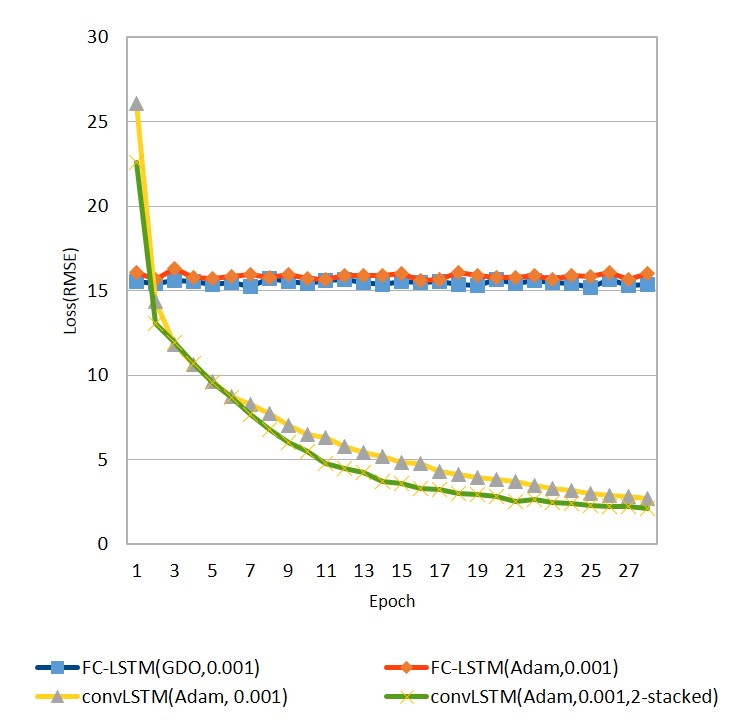}
\caption{Learning curves of differently conditioned models}
\label{figure:learningcurve}
\end{figure}

Figure \ref{figure:learningcurve} shows the learning curves by four different conditions of two models (FC-LSTM and ConvLSTM). ConvLSTM shows significantly better learning performance than FC-LSTM using Adam and Gradient Descendant Optimizer (GDO). A further experiment confirmed that FC-LSTM with GDO required 300 epochs to reduce the loss from 15.5 to 14.4, while ConvLSTM has required only five epochs to reach a loss of 10.0. It seems that the convolution operation efficiently extracted underlying features from the data and enabled quick training. 

\begin{figure}[h]
\centering
\includegraphics[width=0.9\columnwidth]{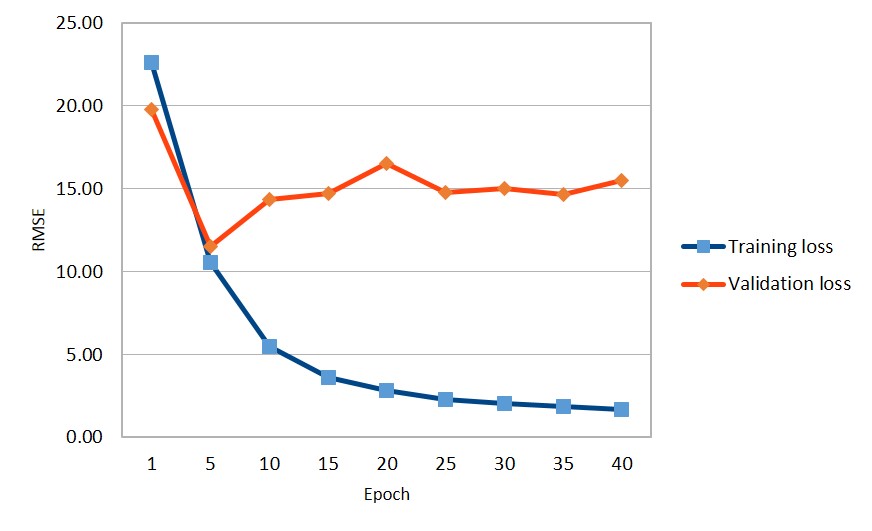}
\caption{Training and validation error curve  with the two-Stacked ConvLSTM}
\label{figure:optEpoch}
\end{figure}

According to Figure \ref{figure:learningcurve}, the two-stacked ConvLSTM model show more stable performance than the one-stacked ConvLSTM. Then, we decided to find the optimal number of epochs to have a predictable performance that was not overfitted with the two-stacked ConvLSTM. Our further experiments with a validation set indicated that the validation loss increases from epoch 5, as shown in Figure \ref{figure:optEpoch}. Then, we measured the performance with the test set from the trained model at epoch 5.

Table \ref{table:rmse} lists the results of measuring the error rate (RMSE) for the test data with multiconditioned models and a baseline. The result of the two-stacked ConvLSTM is shown to be 11.31, which is 23.0\% less than that of the linear regression model used as a control. Furthermore, it is 21.8\% lower than that of the FC-LSTM \cite{kim2017}. The poor performance of the FC-LSTM seems to originate from the fact that the FC-LSTM is fed with one-dimensional input data and loses spatial information in the cell. 

\begin{table}[h]
\centering
\caption{RMSE of predicting rainfall amount with test set}
\label{table:rmse}
\small
\begin{tabular}{p{5.5cm}p{1.0cm}p{1.0cm}}
\hline
Model                   & RMSE & Drop(\%) \\ \hline
Linear Regression&14.69&-  \\
DeepRain: FC-LSTM\cite{kim2017}&14.46&1.6 \\
DeepRain: Conv-LSTM(one-Stacked)&11.51&21.6  \\
DeepRain: Conv-LSTM(two-Stacked)&11.31&23.0   \\\hline
\end{tabular}
\end{table}

\section{Conclusion}
In this study, we first applied ConvLSTM to three-dimensional and four-channel radar data to predict the rainfall amount between 1 h and 2 h. Experimental results showed the prediction accuracy of the proposed methodology is better than that of the linear regression and the FC-LSTM. Future studies will utilize Convolutional Gated Recurrent Units (ConvGRU) to compare ConvLSTM and expand the data set with several data augmentation techniques to enhance the performance. The augmentation technique will include cropping data of a 50 * 50$km^2$ area from the center, which is an important consideration in predicting rainfall. In addition, we are devising an effective convolution method on spatial three-dimensional data with multiple variables and channels. Lastly, we have a plan to consider El Nino for our model, which is likely to have an effect on precipitation of the studied area.


\section*{Acknowledgments}
This work was supported by research projects carried out by the Korea Institute of Science and Technology Information (KISTI): No. K-17-L05-C08, A Research for Typhoon Track Prediction using End-to-End Deep Learning Technique.

\bibliographystyle{ieeetr}
\bibliography{ci_references}

\end{document}